\documentclass[conference]{IEEEtran}
\IEEEoverridecommandlockouts
\usepackage{comment}
\usepackage{hyperref}
\usepackage{cite}
\usepackage{amsmath,amssymb,amsfonts}
\usepackage{algorithmic}
\usepackage{graphicx}
\usepackage{textcomp}
\usepackage{multirow}
\usepackage{authblk}
\usepackage{subcaption}
\usepackage{xcolor}
\def\BibTeX{{\rm B\kern-.05em{\sc i\kern-.025em b}\kern-.08em
    T\kern-.1667em\lower.7ex\hbox{E}\kern-.125emX}}

\newcommand{\orcid}[1]{\href{https://orcid.org/#1}{\includegraphics[width=8pt]{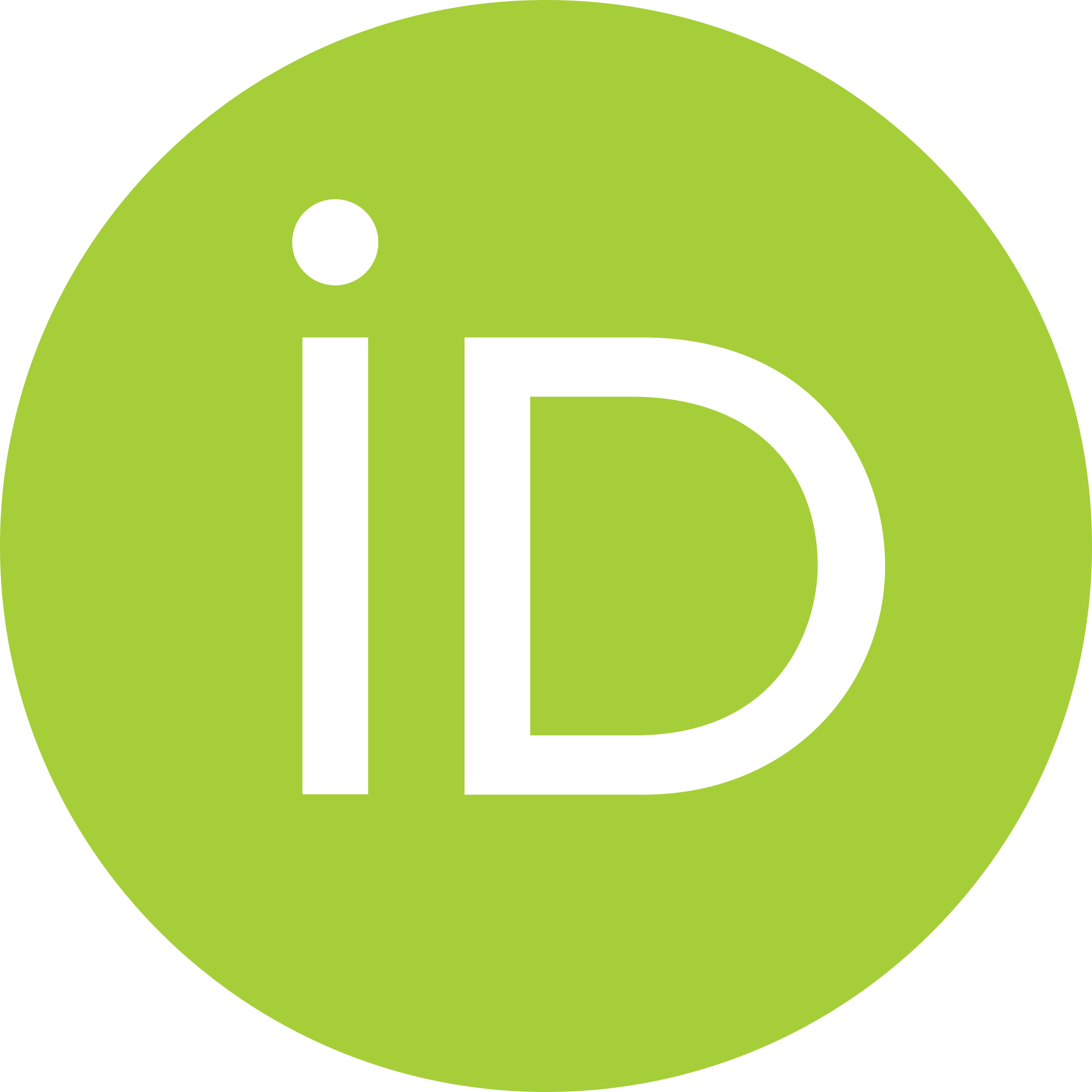}}}

\begin{document}

%\title{A Body-part based Visualisation Framework for the Identification of Abnormal Infants Movements\\}
\title{Towards Explainable Abnormal Infant Movements Identification: A Body-part Based Prediction and Visualisation Framework

\thanks{This project is supported in part by the Royal Society (Ref: IES\slash R1\slash191147).

Corresponding author: Edmond S. L. Ho (e.ho@northumbria.ac.uk)
}
}

\begin{comment}
\author{\IEEEauthorblockN{Kevin D. McCay}
\IEEEauthorblockA{\textit{Computer and Information Sciences} \\
\textit{Northumbria University}\\
Newcastle upon Tyne, UK\\
kevin.d.mccay@northumbria.ac.uk \orcid{0000-0002-3790-1043}}
\and
\IEEEauthorblockN{Edmond S. L. Ho}
\IEEEauthorblockA{\textit{Computer and Information Sciences} \\
\textit{Northumbria University}\\
Newcastle upon Tyne, UK \\
e.ho@northumbria.ac.uk \orcid{0000-0001-5862-106X}}
\and
\IEEEauthorblockN{Dimitrios Sakkos}
\IEEEauthorblockA{\textit{Computer and Information Sciences} \\
\textit{Northumbria University}\\
Newcastle upon Tyne, UK \\
e.ho@northumbria.ac.uk \orcid{0000-0001-5862-106X}}
\and
\IEEEauthorblockN{Claire Marcroft}
\IEEEauthorblockA{\textit{Newcastle Neonatal Service} \\
\textit{NUTH NHS Foundation Trust}\\
Newcastle upon Tyne, UK\\
claire.marcroft@newcastle.ac.uk}
\and
\IEEEauthorblockN{Patricia Dulson}
\IEEEauthorblockA{\textit{Newcastle Neonatal Service} \\
\textit{NUTH NHS Foundation Trust}\\
Newcastle upon Tyne, UK \\
patricia.dulson@nuth.nhs.uk}
\and
\IEEEauthorblockN{Nicholas D. Embleton}
\IEEEauthorblockA{\textit{Newcastle Neonatal Service} \\
\textit{NUTH NHS Foundation Trust}\\
Newcastle upon Tyne, UK\\
nicholas.embleton@newcastle.ac.uk \orcid{0000-0003-3750-5566}}
\and
\IEEEauthorblockN{Wai Lok Woo}
\IEEEauthorblockA{\textit{Computer and Information Sciences} \\
\textit{Northumbria University}\\
Newcastle upon Tyne, UK\\
wailok.woo@northumbria.ac.uk \orcid{0000-0002-8698-7605}}
}
\end{comment}

\author[1]{Kevin D. McCay}
\author[1]{Edmond S. L. Ho}
\author[1]{Dimitrios Sakkos}
\author[1]{Wai Lok Woo}
\author[2]{\\Claire Marcroft}
\author[2]{Patricia Dulson}
\author[2]{Nicholas D. Embleton}
\affil[1]{Department of Computer and Information Sciences, Northumbria University, Newcastle upon Tyne, UK} %\authorcr Email: {\tt \{uid1, uid2\}@usc.edu}\vspace{1.5ex}}
\affil[2]{Newcastle Neonatal Service, NUTH NHS Foundation Trust, Newcastle upon Tyne, UK} %\authorcr Email: {\tt uid3@jpl.nasa.gov} \vspace{-2ex}} 

\maketitle

\begin{abstract}
%Cerebral palsy (CP) is the collective term for a group of lifelong neurological conditions which typically affect movement and coordination. 
Providing early diagnosis of cerebral palsy (CP) is key to enhancing the developmental outcomes for those affected. Diagnostic tools such as the General Movements Assessment (GMA), have produced promising results in early diagnosis, however these manual methods can be laborious. 

%Our previous works established the viability of using pose-based features, extracted from RGB video sequences, for the classification of infant body movements based upon the GMA. 

In this paper, we propose a new framework for the automated classification of infant body movements, based upon the GMA,  which unlike previous methods, also incorporates a visualization framework to aid with interpretability. Our proposed framework segments extracted features to detect the presence of Fidgety Movements (FMs) associated with the GMA spatiotemporally. These features are then used to identify the body-parts with the greatest contribution towards a classification decision and highlight the related body-part segment providing visual feedback to the user.

We quantitatively compare the proposed framework's classification performance with several other methods from the literature and qualitatively evaluate the visualization's veracity. Our experimental results show that the proposed method performs more robustly than comparable techniques in this setting whilst simultaneously providing relevant visual interpretability.

%examine the feasibility of  segmentation-based features by proposing a new image sequence-based classification framework for the diagnosis of cerebral palsy from video data. This new approach allows us to take advantage of additional body-part specific information as a means of improving classification accuracy. This segmentation-based method affords the opportunity to include finer detail than previous methods, as well as providing a framework which c 

\end{abstract}


\section{Introduction}
\label{sec:introduction}

Cerebral palsy (CP) is the term for a group of lifelong neurological conditions which can cause difficulties with mobility, posture and coordination. CP can also cause problems with swallowing, speech articulation, vision, and has been associated with a diminished ability to learn new skills. There is significant variance in the severity of CP, with some individuals showing very minor symptoms whilst others may be severely disabled \cite{nice}. CP is attributed to non-progressive damage to the brain in early infancy \cite{bax_2005, Rosenbaum_CP2007} and is one of the most common physical disabilities in childhood. However, early diagnosis of CP can be difficult, with a confirmed diagnosis rarely made before 18 months of age \cite{Marcroft2015}. The difficulty in providing an early diagnosis is problematic, as early intervention care is considered particularly important for those with emerging and diagnosed CP.  %Early intervention care not only offers a clinical structure which can lessen the impact of impairment \cite{10.1001/jamapediatrics.2017.1689}, but also allows for the appropriate health, social, educational and parental support resources to be put in place \cite{Shevell2013} \cite{Basu2015}.
 
 %Currently, tests which evaluate the complexity, spontaneity, and overall quality of an infant's movements, such as the General Movements Assessment (GMA), are used to identify the emerging signs of CP. The GMA has been developed over a number of years and has proven highly successful in detecting CP \cite{Morgan_2016}, with reports suggesting that the GMA produces more reliable results than other methods, such as cranial ultrasound and neurological examination \cite{Bosanquet:CP_review2013}.
 
 Currently, the General Movements Assessment (GMA) is used to evaluate infant movement by manually observing spontaneous infant movements at a specific stage in development. In a typically developing infant ``Fidgety Movements" (FMs) are detectable from 3 to 5 months post term \cite{Prechtl1997} and consistently have a similar appearance. The absence of these movement characteristics consequently allows for abnormal FM patterns to be identified and classified \cite{Einspieler2016}. However, the challenges associated with applying these assessments in practice depends upon the availability of appropriately trained clinicians. %Given that considerable additional training is required, only infants at high risk of developing CP are currently assessed using the GMA \cite{alliance}. The GMA is also based around gestalt visual perception of movement \cite{EINSPIELER199747}, meaning that quantifiable predictions are difficult to specify.
%These tests are based around the gestalt visual perception of movement \cite{EINSPIELER199747}, and are therefore highly subjective, lacking discernible quantitative diagnostic features. 
%These clinical tests are also heavily reliant upon the infant being in a suitable behavioural state \cite{Fjortoft2009}, making them potentially very time-consuming, which can subsequently lead to observer fatigue.
In order to address the issues surrounding manual clinical assessment, %several works \cite{Adde2010, Stahl2012, Orlandi2018, mccay_pose, McCay_deepPose, Wu:MovCompIdx} have examined the viability of automating the GMA process using machine-learning techniques. %S
several studies have been carried out which attempt to assess the viability of automating assessments to predict motor impairment based upon observed motion quality using computer vision-based approaches \cite{Marcroft2015}. Examples such as \cite{Adde2010} \cite{Adde2013} \cite{Adde2018} explore a per frame background subtraction method for analysis, whereas more recent methods \cite{Stahl2012} \cite{Orlandi2018} %\cite{Ihlen_DL_CP}
\cite{Rahmati_2016}, propose the use of Optical Flow-based methods to track and assess infant movements. Whilst reasonable results are obtained, these methods typically struggle to deal with intra-class variation, as well as anomalies within the recorded video footage such as illumination changes, camera movement, subject-scaling, and resolution inconsistencies. This makes it difficult for these approaches to be adopted in a real-world clinical setting.

On the other hand, with the advancement of pose estimation techniques, high-quality skeletal poses can be extracted from video automatically. Recent work such as \cite{mccay_pose, McCay_deepPose} proposed using histogram-based pose features to automate GMA by classifying infant movements into FM+ (normal) and FM- (abnormal). The pose-based features, namely Histograms of Joint Orientation 2D (HOJO2D) \cite{mccay_pose} and Histograms of Joint Displacement 2D (HOJD2D) \cite{mccay_pose} are computed from the orientation of the body segment and the displacement of the joints, respectively. Encouraging classification performance on traditional classifiers \cite{mccay_pose} and deep learning frameworks \cite{McCay_deepPose} were demonstrated. Wu et al. \cite{Wu:MovCompIdx} proposed {\it Movement Complexity Index} which determines the complexity of the body movements of the infant by computing the correlations between the movements of the joints using the Spearman Correlation Coefficient Matrix (SCCM). Although the method focuses on analyzing the features to predict the risk level of CP of the infant without the need of the training process as in machine learning based approaches, the features are computed from 3D skeletal data which requires specialized image sensing devices to capture those data. Furthermore, the method requires the user to specify a threshold level of the computed index to separate normal/abnormal, and it is unclear if this can be generalized to other datasets. %. Determining such a threshold automatically is not discussed in \cite{Wu:MovCompIdx}. 

The aforementioned studies suggest that an automated system could potentially help to reduce the time and cost associated with current manual clinical assessments, and also assist clinicians in making earlier and more confident diagnoses by providing additional information about the assessed infant movements. However, these methods are also not without their setbacks. One of the main issues with using machine-learning approaches in the medical domain is the problem of interpretable AI. Models are often seen as `black boxes' in which the underlying structures can be difficult to understand. There is an increasing requirement for the mechanisms behind why systems are making decisions to be transparent, understandable and explainable \cite{holzinger2017need}. As such, we propose a new motion classification and visualization framework, which takes an RGB video as the input and analyzes the movement of individual body parts to determine if FMs are present (FM+) or absent (FM-), subsequently identifying normal or abnormal general movements from segments of the sequence. To make our proposed framework fully interpretable, an important aspect is the integration of an automatically generated visualization capable of relaying pertinent information to the assessor. The visualization highlights body-parts which are showing movement abnormalities, and are subsequently providing the most significant contribution towards the classification result. As such, our proposed contributions are summarized as:
\begin{itemize}
    \item A new body-part based classification framework for the automated prediction of CP based upon body movement extracted from videos.
    \item A visualization feature to highlight pertinent body-parts in the video to improve the model interpretability.
\end{itemize}

Experimental results showed that out proposed fidgety movements prediction framework achieved 100\% accuracy and outperforms the existing work on the benchmark MINI-RGBD \cite{hesse2018computer} dataset. The details of our proposed classification and visualisation framework are discussed in Section \ref{sec:Methodology}. Our evaluation is discussed in Section \ref{sec:evaluation}. Our hope is that this contribution will aid in the adoption of such technologies in this domain, through accurate, quantifiable and explainable results. A demo video is available on  \url{https://youtu.be/6CZZmWnT4mo}

\section{Methodology} \label{sec:Methodology}
\begin{figure*}[htb]
\centering
\includegraphics[width=0.85\linewidth]{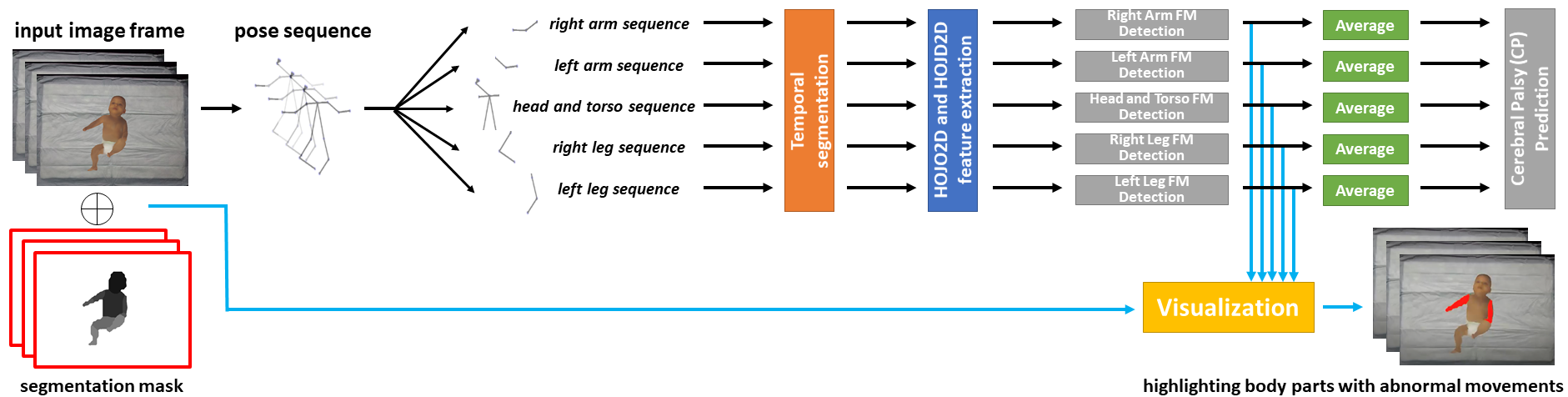}
\caption{The overview of the proposed prediction and visualization framework.}
\label{fig:overview}
\end{figure*}

In this section, we discuss the proposed classification and visualization framework illustrated in Figure \ref{fig:overview}.

\subsection{Pose-based motion features} \label{sec:Pose-feature}
The first step of our proposed framework is extracting features from input video data.  McCay et al. \cite{mccay_pose} demonstrated the effectiveness of using histogram-based motion features, namely Histograms of
Joint Orientation 2D (HOJO2D) and Histograms of Joint
Displacement 2D (HOJD2D), extracted from 2D skeletal poses in detecting FMs from videos. In this paper, an early fusion (i.e. concatenation) of the HOJO2D and HOJD2D is used as the input motion features, since better performance has been demonstrated \cite{mccay_pose,McCay_deepPose}.

\subsection{Spatiotemporal Fidgety Movement Detection} \label{sec:Spatiotemporal}
In order to detect the presence of FMs spatiotemporally, the motion features have to be extracted from 1) different body-parts and 2) different temporal segments individually. Inspired by this, we propose motion feature extraction from 5 different body-parts in the spatial domain, namely {\it left arm, right arm, left leg, right leg, and head-torso}. For the temporal domain, we compute HOJO2D and HOJD2D features (8 bins) for the 5 body parts from every 100-frame segment. In doing so, each video is represented by multiple histogram-based motion features accordingly. For example, a 1000-frame video will be represented by 50 fused features of HOJO2D and HOJD2D.

In this work, we formulate the FMs detection problem as a binary classification. Since each video is annotated with FM+ or FM-, we label all the fused features extracted according to the holistic annotation of the video. When training the classifier all features are used, while the temporal location information is not used. In other words, no matter whether the features are extracted from the beginning or near the end of the video, they will be used to train a single classifier. This proposed approach provides distinct advantages over previous methods, i.e. 1) the classifier will be trained by more data samples rather than using only one histogram representation for the whole video as proposed in \cite{mccay_pose, McCay_deepPose}, and 2) a focus on the presence/absence of FMs while ignoring the temporal information when training the classifier.

We follow McCay et al. \cite{mccay_pose} on using an ensemble classifier on MATLAB R2020a %\cite{MATLAB:2020} 
that consists of a wide range of classifiers to boost the performance of the classification results. Given the multiple fused features extracted from a video, all the features will be classified as FM+ or FM-, this information is then used in visualizing the results (Section \ref{sec:visualization}). As the features were extracted in sequential order in the temporal domain, the classification result on each histogram-based motion feature is essentially detecting FMs spatiotemporally.

\subsection{Late Fusion for Cerebral Palsy Prediction} \label{sec:lateFusion}
While the method presented in Section \ref{sec:Spatiotemporal} provides precise information on the presence/absence of FMs spatiotemporally, directly using all motion features as a cerebral palsy prediction for the whole video will result in sub-optimal performance since the temporal ordering is less important in the GMA than the presence/absence of sustained FMs at any point in the sequence. To tackle this problem, we propose representing each of the 5 body parts using a single scalar score $s$, with this being the average score of the classification result (FM+ as $0$ and FM- as $1$) across all temporal segments for each body-part. Therefore, the range of $s$ will be between $0$ and $1$.

Here, we propose to use a late fusion approach to train an ensemble classifier for cerebral palsy prediction. Specifically, each video is represented by using the 5 scores obtained from the body parts. The binary classifier will predict whether the motion in the video is considered {\it normal} or {\it abnormal}.

\subsection{Visualization} \label{sec:visualization}
While machine learning-based frameworks have obtained excellent performance in a wide range of visual understanding tasks, most of the existing frameworks can be considered black-box approaches since most of the classification frameworks only output the predicted label without specifying exactly what influences the classification decision. Whilst this is acceptable in typical computer vision tasks, it is less preferable in healthcare applications, since it is essential for the clinicians to verify the prediction as well. 

To extract body part information from an input image, the CDCL \cite{lin2019cross} pre-trained body segmentation model is used in this work. The body is segmented into 6 parts; head, torso, upper arms, lower arms, pelvis and upper legs, and lower legs. An example of the segmentation result is illustrated in Figure \ref{fig:overview} (bottom left-hand corner). Specifically, given an input infant image, CDCL \cite{lin2019cross} returns an image mask for segmentation. To align with those 5 body parts to be used in this work, we separate the segmentation masks for the arms and legs into the left and right masks. Here, k-means clustering is used to divide the pixels on each segment mask into two groups. %The mask will be used during the training and testing stages of our proposed framework.

In order to make our proposed framework more interpretable, we include a visualization module that highlights the body-parts that are contributing to the classification decision. Our proposed method highlights the body-parts in {\it red} to indicate the {\it absence of fidgety movements} based on the scores computed in the body part abnormality detection explained in Section \ref{sec:Spatiotemporal}, providing clinicians with an intuitive visualization such as the examples illustrated in Figure \ref{fig:visualization_example}.

\section{Evaluation} \label{sec:evaluation}
In this section, we evaluate the effectiveness of our proposed method using the public dataset MINI-RGBD \cite{hesse2018computer} with Fidgety movement annotation by an experienced GMs assessor in \cite{mccay_pose}. We first compare the performance of our method on fidgety movement detection with baselines methods in Section \ref{sec:res_classification}. Next, we present the visualization results as  qualitative analysis in Section \ref{sec:res_visaulization}. We follow the standard protocol as in  \cite{mccay_pose, McCay_deepPose, Wu:MovCompIdx} to conduct a leave-one-subject out cross-validation to ensure the results presented in this section are obtained base on {\it unseen data} during the training process. % to demonstrate our framework can be generalized to handle

\subsection{Quantitative Evaluation on the Fidgety Movement Detection Results} \label{sec:res_classification}
To demonstrate the overall performance of our proposed framework, we first evaluate the cerebral palsy prediction of the whole input video as explained in Section \ref{sec:lateFusion}. We compared with the existing methods and the results are presented in Table \ref{tabl:fused}. Using our framework, we achieved a perfect prediction with 100\% accuracy. This highlights the effectiveness of our proposed framework over the previous work (\cite{mccay_pose, McCay_deepPose,Wu:MovCompIdx}). %Although McCay et al. \cite{mccay_pose} also achieved 100\% accuracy when LDA is used as the classifier, it can be seen that McCay et al. \cite{mccay_pose}

\begin{table}
\begin{center}
\begin{tabular}{ | c | c | c | c |}
\hline
Method  & {Accuracy}  & {Sensitivity}  & {Specificity}\\
\hline
\textit{\cite{mccay_pose} w/ LDA} & 66.67\% & 50.00\% & 75.00\%\\
\hline
%\textit{\cite{mccay_pose} w/ LDA} & \textbf{100.00\%} & \textbf{100.00\%} & \textbf{100.00\%}\\
%\hline

\textit{\cite{mccay_pose} w/ SVM} & 83.33\% & 50.00\% & \textbf{100.00\%} \\
\hline
%\textit{\cite{mccay_pose} w/ SVM} & 66.67\% & ? & ?  \\
%\hline

\textit{\cite{mccay_pose} w/ Decision Tree}  &  75.00\% & 50.00\% & 87.50\%  \\
\hline
%\textit{Decision Tree}  &  75.00\% & ? & ?  \\
%\hline

\textit{\cite{mccay_pose} w/ kNN (k=1)} &  75.00\% & 25.00\% & {\bf 100.00\%}  \\
\hline
%\textit{\cite{mccay_pose} w/ kNN (k=1)} &  83.33\% & ? & ?  \\
%\hline

\textit{\cite{mccay_pose} w/ kNN (k=3)} &   50.00\%  & 00.00\% & 75.00\%   \\
\hline
%\textit{\cite{mccay_pose} w/ kNN (k=3)} &   83.33\%  & ? & ?   \\
%\hline

\textit{\cite{mccay_pose} w/ Ensemble}  &  66.67\% & 50.00\% & 75.00\%\\
\hline
%\textit{\cite{mccay_pose} w/ Ensemble}  &  75.00\%  & ? & ?  \\
%\hline

\textit{FCNet \cite{McCay_deepPose}} & 83.33\%  & 75\% & 87.5\%   \\
\hline
\textit{Conv1D-1 \cite{McCay_deepPose}} &  83.33\%   & 75\% & 87.5\%  \\
\hline
\textit{Conv1D-2 \cite{McCay_deepPose}}&  91.67\%  & 75\% & {\bf 100.00\%}   \\
\hline
\textit{Conv2D-1 \cite{McCay_deepPose}} &  83.33\%  & 75\% & 87.5\%   \\
\hline
\textit{Conv2D-2 \cite{McCay_deepPose}} & 83.33\% & 75\% & 87.5\%  \\
\hline
\textit{Movement Complexity Index \cite{Wu:MovCompIdx}} & 91.67\% & {\bf 100.00\%} & 87.5\%  \\
\hline
\hline
\textbf{Our method} &  \textbf{100.00\%} &  \textbf{100.00\%} &  \textbf{100.00\%} \\
\hline
%\hline
%\textit{\cite{mccay_pose} w/ LDA} &  \multirow{12}{*}{16} & 83.33\% & ? & ?\\
%\cline{1-1} \cline{3-5}
%\textit{\cite{mccay_pose} w/ SVM} &   &  66.67\% & ? & ? \\
%\cline{1-1} \cline{3-5}
%\textit{\cite{mccay_pose} w/ Decision Tree} &  &  75.00\%  & ? & ? \\
%\cline{1-1} \cline{3-5}
%\textit{\cite{mccay_pose} w/ kNN (k=1)} &   &  75.00\%  & ? & ?  \\
%\cline{1-1} \cline{3-5}
%\textit{\cite{mccay_pose} w/ kNN (k=3)} &  &  66.67\%  & ? & ?  \\
%\cline{1-1} \cline{3-5}
%\textit{\cite{mccay_pose} w/ Ensemble} &  & 75.00\%  & ? & ? \\
%\cline{1-1} \cline{3-5}
%\textit{\cite{McCay_deepPose} w/ FCNet} &   &  83.33\%  & ? & ? \\
%\cline{1-1} \cline{3-5}
%\textit{\cite{McCay_deepPose} w/ Conv1D-1} &   &  83.33\%   & ? & ? \\
%\cline{1-1} \cline{3-5}
%\textit{\cite{McCay_deepPose} w/ Conv1D-2} &  & 91.67\% & ? & ?  \\
%\cline{1-1} \cline{3-5}
%\textit{\cite{McCay_deepPose} w/ Conv2D-1} &   &  75.00\% & ? & ?  \\
%\cline{1-1} \cline{3-5}
%\textit{\cite{McCay_deepPose} w/ Conv2D-2} &   &  83.33\%  & ? & ?  \\
%\cline{1-1} \cline{3-5}
%\textbf{Our method} & &  83.33\% &  80\% &  88.89\% \\
%\hline
\end{tabular}
\end{center}
\caption{ %Fusing the HOJO2D and HOJD2D feature sets: 
Classification accuracy comparison between our proposed framework and baseline methods.}
\label{tabl:fused}
\end{table}

\subsection{Qualitative Evaluation on the Visualization Results} \label{sec:res_visaulization}
\begin{figure}
    \begin{subfigure}[b]{1.0\linewidth}
    \centering
    \includegraphics[trim=3cm 2cm 6cm 2cm, clip, width=0.28\linewidth]{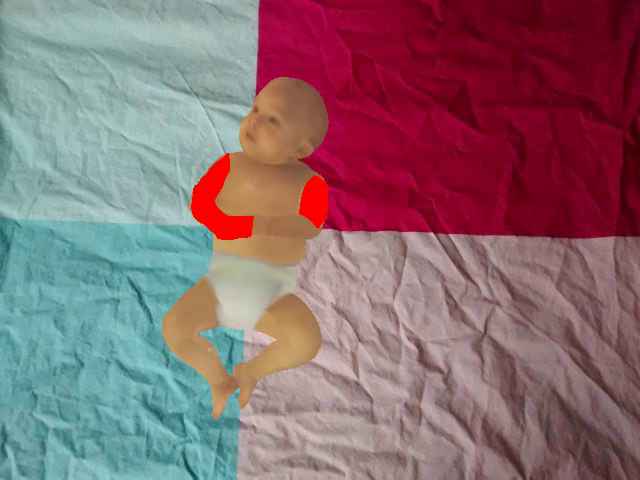}
    \includegraphics[trim=3cm 2cm 6cm 2cm, clip, width=0.28\linewidth]{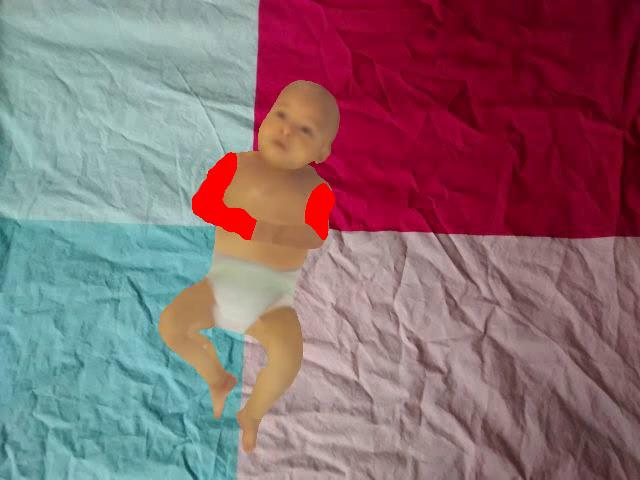}
    \includegraphics[trim=3cm 2cm 6cm 2cm, clip, width=0.28\linewidth]{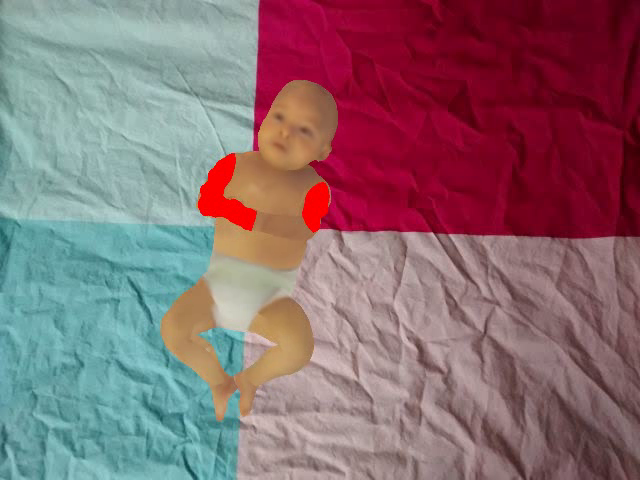}
    \caption{Positive (abnormal) example}
    \end{subfigure} \\
    \begin{subfigure}[b]{1.0\linewidth}
    \centering
    \includegraphics[trim=3cm 2cm 6cm 2cm, clip, width=0.28\linewidth]{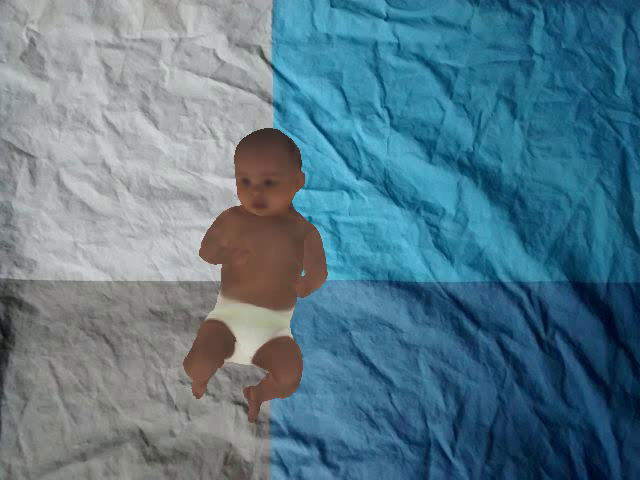}
    \includegraphics[trim=3cm 2cm 6cm 2cm, clip, width=0.28\linewidth]{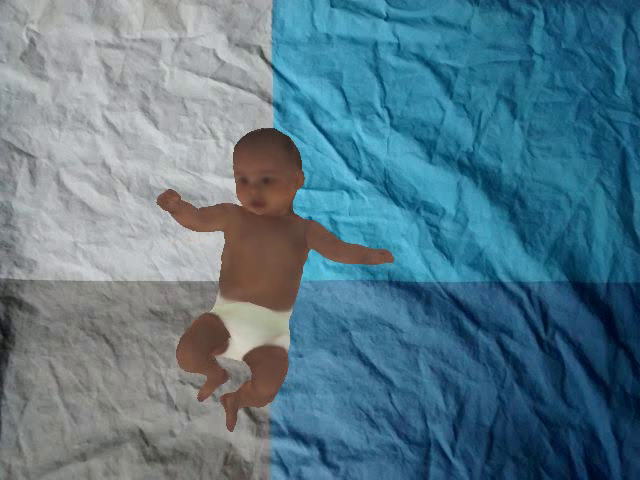}
    \includegraphics[trim=3cm 2cm 6cm 2cm, clip, width=0.28\linewidth]{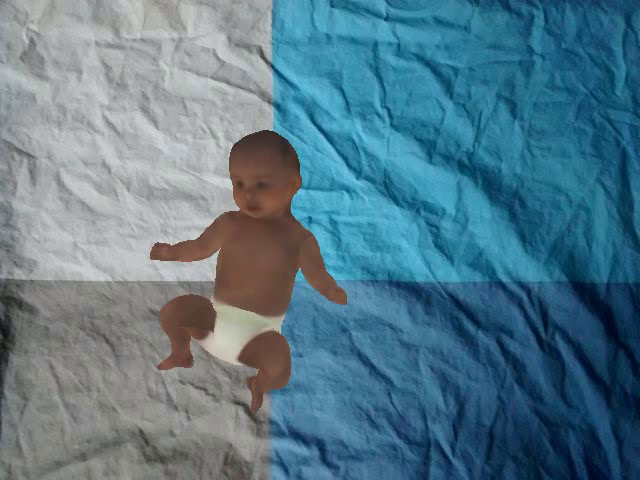}
    \caption{Negative (normal) example}
    \end{subfigure}\\
    \begin{subfigure}[b]{1.0\linewidth}
    \centering
    \includegraphics[trim=3cm 3cm 6cm 1cm, clip, width=0.28\linewidth]{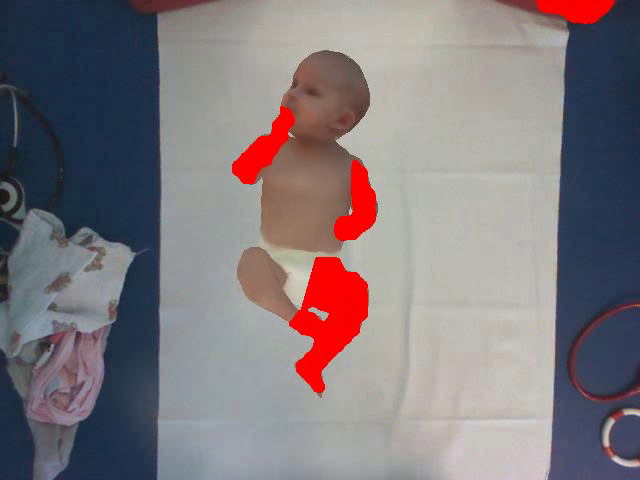}
    \includegraphics[trim=3cm 3cm 6cm 1cm, clip, width=0.28\linewidth]{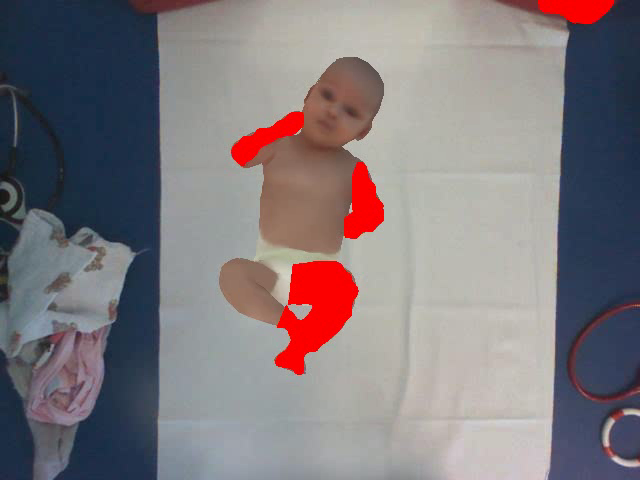}
    \includegraphics[trim=3cm 3cm 6cm 1cm, clip, width=0.28\linewidth]{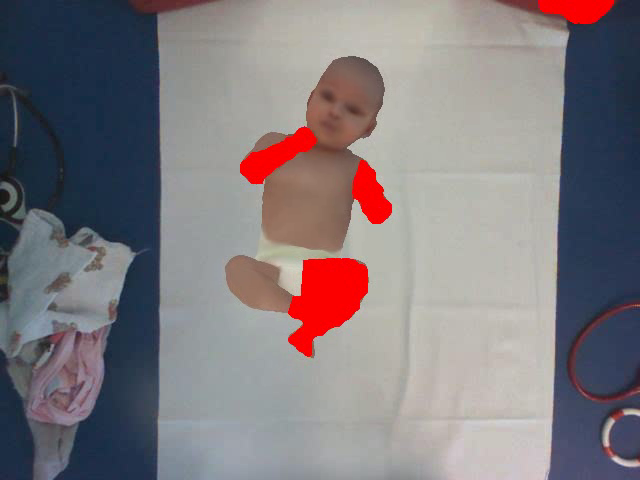}
    \caption{Positive (abnormal) example}
    \end{subfigure}
    \caption{Examples of the video generated by our visualization module. Body-parts without fidgety movements are highlighted in red.}
    \label{fig:visualization_example}
\end{figure}

We further provide qualitative results to demonstrate the effectiveness of our proposed framework. As presented in Section \ref{sec:visualization}, we detect the absence (FM-) or presence (FM+) of fidgety movements of each body part in each temporal segment (see Section \ref{sec:Spatiotemporal}). The body parts with a prediction of FM- will be highlighted in red. An example is illustrated in Figure \ref{fig:visualization_example}. Readers are referred to \url{https://youtu.be/6CZZmWnT4mo} %accompanying video demos 
to evaluate the visual quality of the results. From the results, it can be seen that the highlighted body-parts generally show less complex or more repetitive movements in the videos annotated as FM-. As shown in Figure \ref{fig:visualization_example} (a), the arms are showing a lack of movement and are subsequently predicted as FM- using our framework. On the other hand, the legs are predicted as FM+ as they are showing some movements in that temporal segment. Figure \ref{fig:visualization_example} (c) show an example with monotonous arm and leg movements and our method highlights those body parts as FM- accordingly. For the videos annotated as FM+, such as the example shown in Figure \ref{fig:visualization_example} (b), it can be seen that a much wider variety of movements can be observed. The visualization provides effective visual feedback to the user, as such clinicians can pay greater attention to the highlighted segments for further analysis. 

\section{Conclusion} \label{sec:Conclusion}
%Our study has confirmed that by capitalizing on recent advances in deep learning we are able to successfully model an infant's movement patterns, quantifying CP risk-related FMs. 
In this paper, we present a new framework for detecting fidgety movements of infants spatiotemporally using the pose-based features extracted from RGB videos. Experimental results demonstrated that the new method not only achieves perfect prediction with 100\% accuracy, but also provides the user with visualization on how the machine-learning based framework made the overall prediction relating to the abnormality of the infant's movement. Whilst our system is able to provide rudimentary visual feedback to the user, additional visualization tools would be useful to exploit the extracted spatiotemporal information and provide additional predictive aid to clinicians for this complex diagnostic task. We intend to further implement our method on data gathered in a real-world clinical setting, as well as explore other relevant visualisation methods to provide meaningful and explainable feedback to the user.

%\section*{Acknowledgment}

\bibliographystyle{plain}
\bibliography{references}

\end{document}